\documentclass{article}

\usepackage{setspace}   
\usepackage{lipsum} 

\usepackage{adjustbox}  
\usepackage{xcolor}  
\usepackage{graphicx}  
\graphicspath{ {./figures/} }
\usepackage{float}  
\usepackage{amsmath}
\usepackage{authblk}
\usepackage{booktabs}  
\usepackage{multirow}  
\usepackage{rotating,tabularx}  
 \usepackage{natbib}  

\def\yu#1{{\color{purple}\textbf{[Yu: #1]}}}
\def\gavin#1{{\color{red}\textbf{[Gavin: #1]}}}
\def\junming#1{{\color{blue}\textbf{[Junming: #1]}}}

\def\yu#1{{\color{purple}\textbf{}}}
\def\gavin#1{{\color{red}\textbf{}}}
\def\junming#1{{\color{blue}\textbf{}}}

\usepackage[T1]{fontenc}
\usepackage[utf8]{inputenc}
\usepackage{tgbonum}

\title{Large-scale Quantitative Evidence of Media Impact on Public Opinion toward China}
\author[1]{Junming Huang}
\author[1]{Gavin Cook}
\author[1,*]{Yu Xie}
\affil[1]{Paul and Marcia Wythes Center on Contemporary China and Department of Sociology, Princeton University}
\affil[*]{yuxie@princeton.edu}

\begin{document}

\maketitle

\begin{abstract}
Do mass media influence people's opinion of other countries? Using BERT, a deep neural network-based natural language processing model, this study analyzes a large corpus of 267,907 China-related articles published by The New York Times since 1970. The output from The New York Times is then compared to a longitudinal data set constructed from 101 cross-sectional surveys of the American public's views on China, revealing that the reporting of The New York Times on China in one year explains $54\%$ of the variance in American public opinion on China in the next. This result confirms hypothesized links between media and public opinion and helps shed light on how mass media can influence public opinion of foreign countries.
\end{abstract}


\section*{Introduction}

America and China are the world's two largest economies, and they are currently locked in a tense trade war. In a democratic system, public opinion shapes and constrains political action. How the American public views China thus affects relations between the two countries. Because few Americans have personally visited China, most Americans form their opinions of China and other foreign lands from media depictions. Our paper aims to explain how Americans form their attitudes on China with a case study of how The New York Times may shape public opinion. Our analysis is not causal, but it is informed by a causal understanding of how public opinion may flow from media to citizen. 

There is considerable debate among scholars of public opinion and political communication over how media affect (media is plural) the American mind~\citep{bennett_new_2008}. Many things complicate what might seem like a straightforward relationship. For example, people generally tend to seek out news sources with which they agree~\citep{iyengar_selective_2008}, and politically-active individuals do so more aggressively than the average person~\citep{zaller_nature_1992}. Additionally, with the advent of social media platforms such as Facebook and Twitter, traditional media respond to audience demand more strongly than ever before~\citep{jacobs_informational_2011}.

Most extant survey data indicate that Americans do not seem to like China very much. Many Americans are reported to harbor doubts about China's record on human rights ~\citep{cao_tibet_2015,aldrich_how_2015} and are anxious about China's burgeoning economic, military, and strategic power~\citep{gries_political_2010,yang_china_2012}. They also think that the Chinese political system fails to serve the needs of the Chinese people ~\citep{aldrich_how_2015}. Most Americans, however, recognize a difference between the Chinese state, the Chinese people, and Chinese culture, and they view the latter two more favorably ~\citep{gries_political_2010}. In Fiske’s Stereotype Content Model~\citep{fiske_model_2002}, which expresses common stereotypes as a combination of ``competence'' and ``warmth,'' Asians belong to a set of ``high-status, competitive out-groups'' and rank high in competence but low in warmth~\citep{lin_stereotype_2005}.

The New York Times, which calls itself the ``Newspaper of Record,'' is the most influential newspaper in the USA and possibly even in the Anglophonic world. The digital edition boasts 91 million unique monthly visitors in the US alone ~\citep{the_new_york_times_company_nyt_nodate}, and while the paper’s reach may be impressive, it is yet more significant that the readership of The New York Times represents an elite subset of the American public. Print subscribers to the New York Times have a median household income of \$191,000, three times the median income of US households writ large  ~\citep{johnathan_rothbaum_survey_2019}. Despite the paper's haughty and sometimes condescending reporting, the paper ``has had and still has immense social, political, and economic influence on American and the world''~\cite[page 81]{schwarz_endtimes?:_2014}.

A small body of prior work studies the The New York Times and how The New York Times reports on China. Blood and Phillips uses autoregression methods on time series data to predict public opinion~\citep{blood_recession_1995}. Wu et al. use a similar autoregression technique and find that public sentiment regarding the economy predicts economic performance and that people pay more attention to economic news during recessions~\citep{wu_conditioned_2002}. Peng finds that coverage of China in the paper has been consistently negative but increasingly frequent as China became an economic powerhouse~\citep{peng_representation_2004}. There is very little other scholarship that applies language processing methods to large corpora of articles from The New York Times or other leading papers. Atalay et al. is an exception that uses statistical techniques for parsing natural languages to analyze a corpus of newspaper articles from The New York Times, the Wall Street Journal, and other leading papers in order to investigate the increasing frequency of information technologies in newspaper classifieds~\citep{atalay_new_2018}. 

Our paper aims to advance understanding of how Americans form their attitudes on China with a case study of how The New York Times may shape public opinion. We hypothesize that media coverage of foreign nations affects how Americans view the rest of the world. This model deliberately simplifies the interactions between audience and media and sidesteps many debates. We confine our analysis to a corpus of 267,907 articles on China from The New York Times articles because of the paper's singular influence and importance. We quantify media sentiment with BERT, a state-of-the-art natural language processing model with deep neural networks, and segment sentiment into eight topics to capture the nuance of American opinion toward China. We then use conventional statistical methods to link media sentiment to a longitudinal data set constructed from 101 cross-sectional surveys of the American public's views on China. We find strong correlations between how The New York Times reports on China in one year and the views of the public on China in the next. The correlations agree with our hypothesis and imply a strong connection between media sentiment and public opinion.

\section*{Results}

We begin with a demonstration of how the reporting of The New York Times on China changes over time, and we follow this with an analysis of how coverage of China might influence public opinion toward China. 

\subsection*{Quantifying media sentiment and public opinion}

We quantify \textbf{media sentiment} with a natural language model on a large-scale corpus of 267,907 articles on China from The New York Times published between 1970 and 2019. To explore sentiment from this corpus in greater detail, we map every article to a sentiment category (positive, negative, or neutral) in eight topics: ideology, government \& administration, democracy, economic development, marketization, welfare and well-being, globalization, and culture. 

We do this with a three-stage modeling procedure. First, two human coders annotate 873 randomly selected articles with a total of 18,598 sentences as expressing either positive, negative, or neutral sentiment in each topic. We treat irrelevant articles as neutral sentiment. Secondly, we fine-tune a natural language processing model BERT (Bidirectional Encoder Representations from Transformers~\citep{bert}) with the human-coded labels. The model uses a deep neural network with 12 layers. It accepts sentences (i.e., word sequences of no more than 128 words) as input and outputs a probability for each category. We end up with two binary classifiers for each topic for a grand total of 16 classifiers: an \textit{assignment} classifier that determines whether a sentence expresses sentiment in a given topic domain and a \textit{sentiment} classifier that then distinguishes positive and negative sentiment in a sentence classified as belonging to a given topic domain. Thirdly, we run the 16 trained classifiers on each sentence in our corpus and assign category probabilities to every sentence. We then use the probabilities of all the sentences in an article to determine the article's overall sentiment category (i.e., positive, negative, or neutral) in every topic.

As demonstrated in Table~\ref{tab:106-accuracy}, the two classifiers are accurate at both the sentence level and the article level. The assignment classifier and the sentiment classifier reach classification accuracy of 89 - 96\% and 73 - 90\% respectively on sentences. The combined outcome of the classifiers, namely article sentiment, is accurate to 65 - 95\% across all eight topics. For comparison, a random guess would reach an accuracy of 50\% on each task (See Supplementary Information for details).

\textbf{American public opinion} towards China is a composite measure drawn from national surveys that ask respondents for their opinions on China. We collect 101 cross-sectional surveys from 1974 to 2019 that asked relevant questions about attitudes toward China and develop a probabilistic model to harmonize different survey series with different scales (e.g., 4 levels, 10 levels) into a single time series, capitalizing on "seaming" years in which different survey series overlapped. For every year, there is a single real value representing American sentiment on China relative to the level in 1974. Put another way, we use sentiment in 1974 as a baseline measure to normalize the rest of the time series. A positive value shows a more favorable attitude than that in 1974, and a negative value represents a less favorable attitude than that in 1974. Because of this, the trends in sentiment changes year-over-year are of interest, but the absolute values of sentiment in a given year are not (See Supplementary Information for details). As shown in Figure~\ref{fig:survey}, public opinion towards China has varied greatly from 1974 to 2019. It steadily climbed from a low of -24\% in 1976 to a high of 73\% in 1987, and has fluctuated between 10\% to 48\% in the intervening 30 years.

\subsection*{Trend of media sentiment}
%
The New York Times has maintained steady interest in China and China-adjacent topics over the years in our sample and has published at least 3,000 articles on China in every year of our corpus. Figure~\ref{fig:nyt-volume-trend} displays the yearly volume of China-related articles from The New York Times on each of the eight topics since 1970. Articles on China increase sharply after 2000 and eventually reach a peak around 2010 at almost double their volume from the 1970s. 
As the number of articles on China increases, the amount of of attention paid to each of the eight topics diverges. Articles on government, democracy, globalization, and culture are consistently common while articles on ideology are consistently rare. In contrast, articles on China's economy, marketization, and welfare were rare before 1990 but become increasingly common after 2000. The timing of this uptick coincides neatly with worldwide recognition of China's precipitous economic ascent and specifically the beginnings of China's talks to join the World Trade Organization.

While the proportion of articles in each given topic change over time, the sentiment of articles in each topic is remarkably consistent. Ignoring neutral articles, Figure~\ref{fig:nyt-sentiment-trend} illustrates the yearly fractions of positive and negative articles about each of the eight topics. We find four topics (economics, globalization, culture, and marketization) are almost always covered positively while reporting on the other four topics (ideology, government \& administration, democracy, and welfare \& well-being) is overwhelmingly negative.

The NYT views China's globalization in a very positive light. Almost 100\% of the articles mentioning this topic are positive for all of the years in our sample. This implies that New York Times welcomes China's openness to the world and, more broadly, may be particularly partial to globalization in general. 

Similarly, economics, marketization, and culture are covered most commonly in positive tones that have only grown more glowing over time. Positive articles on these topics begin in 1970s when China and the US began Ping-Pong diplomacy, and eventually comprise $1/4$ to $1/2$ of articles in these three topics, the remainder of which are mostly neutral articles. This agrees with the intuition that most Americans like Chinese culture. The New York Times has been deeply enamored with Chinese cultural products ranging from Chinese art to Chinese food since the very beginning of our sample. Following China's economic reforms, the number of positive articles and the proportion of positive articles relative to negative articles increases for both economics and marketization.

In contrast, welfare \& well-being is covered in an almost exclusively negative light. About $1/4$ of the articles on this topic are negative, and almost no articles on this topic are positive. Topics regarding or adjacent to politics are covered very negatively. Negative articles on ideology, government \& administration, and democracy outnumber positive articles on these topics for all of the years in our sample. Though small fluctuations that coincided with ebbs in US-China relations are observed for those three topics, coverage has only grown more negative over time. Government \& administration is the only negatively-covered topic that does feature some positive articles. This reflects the qualitative understanding that The New York Times thinks that the Chinese state is evil but capable.

Despite the remarkable diversity of sentiment toward China across the eight topics, sentiment within each of the topics is startlingly consistent over time. This consistency attests to the incredible stability of American stereotypes towards China. If there is any trend to be found here, it is that the main direction of sentiment in each topic, positive or negative, has grown more prevalent since the 1970s. This is to say that reporting on China has become more polarized, which is reflective of broader trends of media polarization ~\citep{jacobs_informational_2011, mullainathan_market_2005}.

\subsection*{Media sentiment affects public opinion}
To reveal the connection between media sentiment and public opinion, we run a linear regression model (Equ~\ref{equ:media2attitude}) to fit public opinion with media sentiment from current and preceding years. 
\begin{equation}
\label{equ:media2attitude}
    \mu_t = \sum_{1 \leq k \leq 8} \sum_{j \in [t, t-1, t-2, \cdots]} \sum_{s \in \{positive, negative\}} \beta_{kjs} F_{kjs},
\end{equation}
where $\mu_t$ denotes public opinion in year $t$ with possible values ranging from $-1$ to $1$. $F_{kjs}$ is the fraction of positive ($s=positive$) or negative ($s=negative$) articles on topic $k$ in year $j$. A non-negative coefficient $\beta_{kjs}$ quantifies the importance of $F_{kjs}$ in predicting $\mu_t$.

There is inertia to public opinion. A broadly-held opinion is hard to change in the short term, and it may require a while for media sentiment to affect how the public views a given issue. For this reason, $j$ is allowed to take $[t, t-1, t-2, \cdots]$ anywhere from zero to a couple of years ahead of $t$. In other words, we inspect lagged values of media sentiment as candidate predictors for public attitudes towards China. 

We seek an optimal solution of media sentiment predictors to explain the largest fraction of variance ($r^2$) of public opinion. To reduce the risk of overfitting, we first constrain the coefficients to be non-negative after reverse-coding negative sentiment variables, which means we assume that positive articles have either no impact or positive impact and that negative articles have either zero or negative impact on public opinion. Secondly, we require that the solution be sparse and contain no more than one non-zero coefficient in each topic:
\begin{equation*}
\begin{aligned}
& \underset{\beta}{\text{maximize}} & & r^2(\mu, \beta, F) \\
& \text{subject to} & & \beta_{kjs} \geq 0, \forall k, j, s \\
&                   & & \| \beta_{k,\cdot,\cdot}\|_0 \leq 1, \forall k \\
\end{aligned}
\end{equation*}
where $r^2(\mu, \beta, F)$ is the explained variance of $\mu$ fitted with $(\beta, F)$. The $l_0$-norm $\| \beta_{k,\cdot,\cdot}\|_0$ gives the number of non-zero coefficients of topic $k$ predictors.

The solution varies with the number of topics we let the model use for fitting. As shown in Table~\ref{tab:201-feature-selection}, if we allow fitting with only one topic, we find that sentiment on Chinese culture has the most explanatory power, accounting for $31.2\%$ of the variance in public opinion. We run a greedy strategy to add additional topics that yield the greatest increase in explanatory power, resulting in eight nested models (Table~\ref{tab:201-feature-selection}). The explanatory power of our models increases monotonically with the number of allowed topics but reaches a saturation point at which the marginal increase in variance explained per topics decreases after only two topics are introduced (See Table~\ref{tab:201-feature-selection}). To strike a balance between simplicity and explanatory power, we use the top two predictors, which are the positive sentiment of culture and the negative sentiment of democracy in the previous year, to build a linear predictor of public opinion that can be written as:
\begin{equation}
\label{equ:media-predict-attitude}
    \mu_t = -0.791 + 3.112 F_{culture, t-1, positive} + 1.452 F_{democracy, t-1, negative}, 
\end{equation}
where $F_{culture, t-1, positive}$ is the yearly fraction of positive articles on Chinese culture in year $t-1$ and $F_{democracy, t-1, negative}$ is the yearly fraction of negative articles on Chinese democracy in year $t-1$. This formula explains $53.9\%$ of the variance of public opinion in the time series. For example, in 1993 $53.9\%$ of the articles on culture had positive sentiment, and $44.2\%$ of the articles on democracy had negative sentiment ($F_{culture, 1993, positive} = 0.539, F_{democracy, 1993, negative} = -0.442$). Substituting those numbers into Equ~\ref{equ:media-predict-attitude} predicts public opinion in the next year (1994) to be $0.236$, very close to the actual level of public opinion ($0.218$) (Figure~\ref{fig:survey-pred}). 

\section*{Discussion}

By analyzing a corpus 267,907 articles from The New York Times with BERT, a state-of-the-art natural language processing model, we identify major shifts in media sentiment towards China across eight topic domains over 50 years and find that media sentiment leads public opinion. Our results show that the reporting of The New York Times on culture and democracy in one year explains 53.9\% of the variation in public opinion on China in the next. 
Our analysis is neither conclusive nor causal, but it is suggestive. The story that we draw from our results is that elite media sentiment on China predicts public opinion on China. There are a number of potential factors that may complicate this story. We do not consider intermediary processes through which opinion from elite media percolates to the masses below. Our results are best interpreted as a ``reduced-form'' description of the overall relationship between media sentiment and public opinion towards China. Though The New York Times might be biased, and it might have a particular bias to how it covers China, the paper's ideological slant does not affect our work explaining the trends in public opinion of China as long as the relevant biases are consistent over the time period covered by our analyses. 

In addition to those specified above, a number of possible extensions of our work remain ripe targets for further research. Though a fully causal model of our text analysis pipeline may prove elusive~\citep{2018-egami2018}, future work may use randomized vignettes to further our understanding of the causal effects of the exposure to media on attitudes towards China. 
Secondly, our modeling framework is deliberately simplified. The state affects news coverage before news ever makes its way to the citizenry. It is plausible that multiple state-level actors may bypass the media and alter public opinion directly and to different ends. For example, the actions and opinions of individual US high-profiled politicians may attenuate or exaggerate the impact of state-level tension on public sentiment toward China. There are presumably a whole host of intermediary processes through which opinion from elite media on high percolates to the masses below. Thirdly, the relationship between the sentiment of The New York Times and public opinion may be very different for hot-button social issues of first-line importance in the American culture wars. In our corpus, The New York Times has covered globalization almost entirely positively, but the 2016 election of President Donald J. Trump suggests that many Americans do not share the zeal of The Times for international commerce. We also plan to extend our measure of media sentiment to include text from other newspapers. The Guardian, a similarly elite, Anglophonic, and left-leaning paper, will make for a useful comparison case. Finally, our analysis was launched in the midst of heightened tensions between the US and China and concluded right before the outbreak of a global pandemic. Many things have changed since COVID-19. Returning to our analysis with an additional year or two of data will almost certainly provide new results of additional interest. 

\bibliographystyle{agsm} 



\section*{Data availability}
\label{section:data-availability}
All data analyzed during the current study are publicly available.
\label{data:nyt}
The New York Times data were accessed using official online APIs (https://developer.nytimes.com/). We use query API to search for 267,907 articles from The New York Times that mention China, Chinese, Beijing, Peking, or Shanghai. We download the full text and date of each article for natural language processing.
The survey data were obtained from three large public archives, namely Roper Center for Public Opinion Research (ROPER), the National Opinion Research Center (NORC), and Inter-university Consortium for Political and Social Research (ICPSR).
The source codes and pretrained parameters of the natural language processing model (BERT) were publicly released by Google Inc. (https://github.com/google-research/bert). 

\newpage

\section*{Figures}

\begin{figure}[ht]
\centering
\caption{\textbf{Public opinion of Americans toward China.} This time series is aggregated from 101 cross-sectional surveys from 1974 to 2019 that asked relevant questions about attitudes toward China, ranging from -100\% to 100\% with the year of 1974 as baseline.
Years with attitudes above zero show a more favorable attitude than that in 1974, with a peak of 73\% in 1987.
Years with attitudes below zero show a less favorable attitude than that in 1974, with a lowest level of -24\% in 1976.
The time series is shown with 95\% confidence interval.
}
\label{fig:survey}
\end{figure}

\begin{figure}[ht]
\caption{\textbf{Topic-specific yearly volume of The New York Times articles with sentiment on China from 1970 to 2019.} In each year we report in each topic the number of positive and negative articles, while ignoring neutral/irrelevant articles. The media have consistently high attention on reporting China government \& administration, democracy, globalization, and culture. There are emerging interests on China's economics, marketization, and welfare \& well-being since 1990s. Note that the sum of the stacks does not equal to the total volume of articles about China, because each article may express sentiment in none or multiple topics.} 
\label{fig:nyt-volume-trend}
\end{figure}

\begin{figure}[ht]
\caption{\textbf{Sentiments on The New York Times on China in eight topics from 1970 to 2019.} 
Each panel reports the trend of yearly media attitude toward China in each of eight domains since 1970. 
The media attitude is measured as the percentages of positive articles and negative articles respectively. 
US-China relation milestones are marked as gray dots.
New York Times express diverging but consistent attitudes in the eight domains, with negative articles consistently common in ideology, government, democracy, and welfare, and positive sentiments common in economic, globalization and culture. Standard errors are too small to be visible (below 1.55\% in all topics all years).}
\label{fig:nyt-sentiment-trend}
\end{figure}

\begin{figure}[ht]
\centering
\caption{\textbf{Regressing public opinion of Americans toward China on The New York Times sentiments.} The public opinion (solid), as a time series, is well fitted by the media sentiments on two selected topics, namely Culture and Democracy, in the previous year. The dash line shows a linear prediction based on the fractions of positive articles on Culture and negative articles on Democracy in the previous year. The public opinion is shown with 95\% confidence interval, and the fitted line is shown with one standard error.}
\label{fig:survey-pred}
\end{figure}

\newpage

\section*{Tables}

\begin{table}[ht]
\caption{\textbf{Accurately quantifying media sentiment.} Media article sentiment accuracy is measured by the percentage of correctly labeled articles (positive, negative, neutral/irrelevant) in each topic. Sentence assignment accuracy is measured as AUC (Area under ROC curve) on each sentence by a binary classifier (positive/negative vs neutral/irrelevant). Sentence sentiment accuracy is measured AUC on each sentence by a binary classifier (positive vs negative).}
\label{tab:106-accuracy}
\begin{tabular}{l|l|l|l}
\toprule
{} & Article & Sentence & Sentence \\
{} & sentiment & assignment & sentiment \\\hline
\midrule
\textbf{Ideology} & $88.7 \pm 3.9 \%$ & $94.1 \pm 1.9 \%$ & $90.4 \pm 7.4 \%$ \\\hline 
\textbf{Government} & $62.7 \pm 6.6 \%$ & $95.6 \pm 0.8 \%$ & $78.1 \pm 4.3 \%$ \\ 
\textbf{Administration}  &  &  &  \\\hline 
\textbf{Democracy} & $87.7 \pm 5.2 \%$ & $95.7 \pm 1.1 \%$ & $77.5 \pm 5.4 \%$ \\\hline 
\textbf{Economic development} & $81.2 \pm 6.3 \%$ & $94.9 \pm 1.1 \%$ & $79.9 \pm 5.0 \%$ \\\hline 
\textbf{Marketization} & $91.3 \pm 4.3 \%$ & $96.9 \pm 1.7 \%$ & $74.9 \pm 14.5 \%$ \\\hline 
\textbf{Welfare} & $82.0 \pm 4.4 \%$ & $94.5 \pm 2.7 \%$ & $73.2 \pm 9.8 \%$ \\ 
\textbf{well-being}  &  &  &  \\\hline 
\textbf{Globalization} & $77.4 \pm 5.8 \%$ & $93.3 \pm 1.5 \%$ & $80.7 \pm 5.6 \%$ \\\hline 
\textbf{Culture} & $79.8 \pm 4.8 \%$ & $90.0 \pm 1.9 \%$ & $73.5 \pm 5.4 \%$ \\\hline 
\bottomrule
\end{tabular}
\end{table}

\begin{sidewaystable}
\caption{\textbf{Nested models fitting public opinion with media sentiment.} We regress public opinion on various numbers of media sentiment predictors, requiring each topic with (a) no more than one predictor, and (b) non-negative coefficients. The best two topic predictors (yearly fraction of positive articles on Chinese culture in the previous year, and yearly fraction of negative articles on Chinese democracy in the previous year) explain $53.9\%$ of the variation in public opinion.}
\label{tab:201-feature-selection}
\begin{tabular}{l|l|l|l|l|l|l|l|l}
\toprule
{} & 1 topic & 2 topics & 3 topics & 4 topics & 5 topics & 6 topics & 7 topics & 8 topics \\
\midrule
$F_{culture, t-1, positive}$ & 2.479 & 3.112 & 2.233 & 2.044 & 1.826 & 1.551 & 1.082 & 1.070\\ 
$F_{democracy, t-1, negative}$ &  & 1.452 & 1.498 & 1.290 & 1.196 & 1.115 & 1.188 & 1.067\\ 
$F_{globalization, t-1, positive}$ &  &  & 1.945 & 1.951 & 2.047 & 1.926 & 1.963 & 1.818\\ 
$F_{government \&  administration, t-4, negative}$ &  &  &  & 0.762 & 1.087 & 1.197 & 1.246 & 1.249\\ 
$F_{marketization, t-4, negative}$ &  &  &  &  & 2.869 & 3.816 & 2.791 & 3.186\\ 
$F_{welfare \&  well-being, t-5, positive}$ &  &  &  &  &  & 5.412 & 8.245 & 8.297\\ 
$F_{economic  development, t-2, negative}$ &  &  &  &  &  &  & 1.607 & 1.794\\ 
$F_{ideology, t-2, negative}$ &  &  &  &  &  &  &  & 0.324\\ 
Intercept & -1.077 & -0.791 & -2.136 & -1.638 & -1.401 & -1.146 & -0.846 & -0.692\\\hline 
Explained variance & 0.312 & 0.539 & 0.577 & 0.606 & 0.622 & 0.639 & 0.656 & 0.659\\\hline 
\bottomrule
\end{tabular}
\end{sidewaystable}

\newpage

\section*{Competing interests}
The authors declare no competing interests.


\section*{Acknowledgments}
The Acknowledgments are hidden in the review phase to satisfy the requirement of author anonymisation.

\end{document}